\pdfoutput=1

\documentclass[11pt]{article}

\usepackage{acl}

\usepackage{times}
\usepackage{latexsym}
\usepackage{natbib}
\usepackage{hyperref}

\usepackage[T1]{fontenc}

\usepackage[utf8]{inputenc}

\usepackage{microtype}
\usepackage{multirow}
\usepackage{graphicx}

\title{Every word counts: A multilingual analysis of individual human alignment with model attention}

\author{Stephanie Brandl \\
  Department of Computer Science \\
  University of Copenhagen \\
  \texttt{brandl@di.ku.dk} \\\And
  Nora Hollenstein \\
  Center for Language Technology \\
  University of Copenhagen \\
  \texttt{nora.hollenstein@hum.ku.dk} \\}

\begin{document}
\maketitle
\begin{abstract}
Human fixation patterns have been shown to correlate strongly with Transformer-based attention. Those correlation analyses are usually carried out without taking into account individual differences between participants and are mostly done on monolingual datasets making it difficult to generalise findings. In this paper, we analyse eye-tracking data from speakers of 13 different languages reading both in their native language (L1) and in English as language learners (L2). We find considerable differences between languages but also that individual reading behaviour such as skipping rate, total reading time and vocabulary knowledge (LexTALE) influence the alignment between humans and models to an extent that should be considered in future studies.
\end{abstract}

\section{Introduction}
Recent research has shown that relative importance metrics in neural language models correlate strongly with human attention, i.e., fixation durations extracted from eye-tracking recordings during reading \citep{morger-etal-2022-cross, eberle-etal-2022-transformer, bensemann-etal-2022-eye,hollenstein-beinborn-2021-relative,sood-etal-2020-interpreting}. This approach serves as an interpretability tool and helps to quantify the cognitive plausibility of language models. 
However, what drives these correlations in terms of differences between individual readers has not been investigated.

In this short paper, we approach this by analysing (i) differences in correlation between machine attention and human relative fixation duration across languages, (ii) differences within the same language across datasets, text domains and native speakers of different languages, (iii) differences between native speakers (L1) and second language learners (L2), (iv) the influence of syntactic properties such as part-of-speech tags, and (v) the influence of individual differences in demographics, i.e., age, vocabulary knowledge, depth of processing.

Taking into account individual and subgroup differences in future research, will encourage single-subject and cross-subject evaluation scenarios which will not only improve the generalization capabilities of ML models but also allow for adaptable and personalized technologies, including applications in language learning, reading development or assistive communication technology. Additionally, understanding computational language models from the perspectives of different user groups can lead to increased fairness and transparency in NLP applications.

\paragraph{Contributions} We quantify the individual differences in human alignment with Transformer-based attention in a correlation study where we compare relative fixation duration from native speakers of 13 different languages on the MECO corpus \citep{meco_corpus,kuperman2022text} to first layer attention extracted from mBERT \cite{devlin-etal-2019-bert}, XLM-R \cite{conneau-etal-2020-unsupervised} and mT5 \cite{xue-etal-2021-mt5}, pre-trained multilingual language models. We carry out this correlation analysis on the participants' respective native languages (L1) and data from an English experiment (L2) of the same participants. We analyse the influence of processing depth, i.e., quantifying the thoroughness of reading through the readers' skipping behaviour, part-of-speech (POS) tags, and vocabulary knowledge in the form of LexTALE scores on the correlation values. Finally, we compare correlations to data from the GECO corpus, which contains English (L1 and L2) and Dutch (L1) eye-tracking data \citep{cop2017presenting}.

The results show that (i) the correlation varies greatly across languages, (ii) L1 reading data correlates less with neural attention than L2 data, (iii) generally, in-depth reading leads to higher correlation than shallow processing. Our code is available at \href{https://github.com/stephaniebrandl/eyetracking-subgroups}{\nolinkurl{github.com/stephaniebrandl/eyetracking-subgroups}}.
\section{Related Work}

\paragraph{Multilingual eye-tracking}
\citet{brysbaert2019many} found differences in word per minute rates during reading across different languages and proficiency levels. That eye-tracking data contains language-specific information is also concluded by \citet{berzak-etal-2017-predicting}, who showed that eye-tracking features can be used to determine a reader’s native language based on English text. 

\paragraph{Individual differences}
The neglection of individual differences is a well-known issue in cognitive science, which leads to theories that support a misleading picture of an idealised human cognition that is largely invariant across individuals \citep{levinson2012original}. 
\citet{kidd2018individual} pointed out that the extent to which human sentence processing is affected by individual differences is most likely underestimated since psycholinguistic experiments almost exclusively focus on a homogeneous sub-sample of the human population \citep{henrich2010weirdest}.

Along the same lines, when using cognitive signals in NLP, most often the data is aggregated across all participants \citep{hollenstein-etal-2020-towards,klerke-plank-2019-glance}. While there is some evidence showing that this leads to more robust results regarding model performance, it also disregards differences between subgroups of readers.

\paragraph{Eye-tracking prediction and correlation in NLP}
State-of-the-art word embeddings are highly correlated with eye-tracking metrics \cite{hollenstein-etal-2019-cognival, salicchi-etal-2021-looking}. \citet{hollenstein-etal-2021-multilingual} showed that multilingual models can predict a range of eye-tracking features across different languages. This implies that Transformer language models are able to extract cognitive processing information from human signals in a supervised way. 
Moreover, relative importance metrics in neural language models correlate strongly with human attention, i.e., fixation durations extracted from eye-tracking recordings during reading \citep{morger-etal-2022-cross, eberle-etal-2022-transformer, bensemann-etal-2022-eye,hollenstein-beinborn-2021-relative,sood-etal-2020-interpreting}.

\setlength{\tabcolsep}{4pt} 
\begin{table*}[t]
\begin{tabular}{l|l|lllllllllllll||ll}
& & \multicolumn{13}{c||}{MECO} & \multicolumn{2}{c}{GECO}\\ 
&  &   de &    el &    en &    es &    et &    fi &    he &    it &    ko &    nl &    no &    ru &    tr & en & nl \\ \hline
\parbox[t]{2mm}{\multirow{3}{*}{\rotatebox[origin=c]{90}{L1}}}
 & mBERT &  0.45 &  0.57 &  0.27 &  0.42 &  0.52 &  0.51 &  0.49 &  0.35 &  0.45 &  0.38 &  0.41 &  0.53 &  0.48 & 0.26 & 0.26\\ 
& XLM-R &  0.53 &  0.66 &  0.37 &  0.54 &  0.6 &  0.59 &  0.55 &  0.47 &  0.51 &  0.48 &  0.52 &  0.65 &  0.53 & 0.27 & 0.28 \\
& mT5 &  0.31 &  0.45 &  0.11 &  0.24 &  0.37 &  0.36 &  0.27 &  0.16 &  0.35 &  0.27 &  0.23 &  0.3 &  0.23 & 0.16 & 0.23\\ \hline
\parbox[t]{2mm}{\multirow{3}{*}{\rotatebox[origin=c]{90}{L2}}}
& mBERT & 0.32 &  0.33 &  0.26 &  0.32 &  0.32 &  0.32 &  0.33 &  0.34 & - & 0.3 &  0.31 &  0.33 &  0.33 & - & 0.29 \\
& XLM-R & 0.42 &  0.43 &  0.35 &  0.41 &  0.42 &  0.42 &  0.42 &  0.45 & - & 0.39 &  0.4 &  0.42 &  0.43 & - & 0.29 \\
& mT5 &  0.11 &  0.13 &  0.08 &  0.12 &  0.13 &  0.13 &  0.12 &  0.13 & - &  0.11 &  0.11 &  0.13 &  0.13 & - & 0.18 \\
\end{tabular}
\caption{Spearman correlation between first layer attention and total reading time for each language and different models.\textsuperscript{\ref{languages}} Correlation values are calculated individually per participant and sentence and averaged across both afterwards. First 3 row show results for L1 languages and the remaining rows show results for the same participants on the L2 English reading task. English L2 data for Korean (ko) participants in MECO and English L2 participants in GECO is not available.}
\label{tab:lang}
\end{table*}

\section{Method}
We analyse the Spearman correlation coefficients between first layer attention in a multilingual language model and relative fixation durations extracted from a large multilingual eye-tracking corpus, including 13 languages \citep{meco_corpus,kuperman2022text} as described below.

Total fixation time (TRT) per word is divided by the sum over all TRTs in the respective sentence to compute relative fixation duration for individual participants, similar to \citet{hollenstein-beinborn-2021-relative}.

We extract first layer attention for each word from mBERT\footnote{\url{https://huggingface.co/bert-base-multilingual-cased}}, XLM-R\footnote{\url{https://huggingface.co/xlm-roberta-base}} and mT5\footnote{\url{https://huggingface.co/google/mt5-base}}, all three are multilingual pre-trained language models. We then average across heads. We also test gradient-based saliency and attention flow, which show similar correlations but require substantially higher computational cost. This is in line with findings in \citet{morger-etal-2022-cross}.

\paragraph{Eye-tracking Data} The L1 part of the MECO corpus contains data from native speakers reading 12 short encyclopedic-style texts (89-120 sentences) in their own languages\footnote{\label{languages}The languages in MECO L1 include: Dutch (nl), English (en), Estonian (et), Finnish (fi), German (de),  Greek (el), Hebrew (he), Italian (it), Korean (ko), Norwegian (no), Russian (ru), Spanish (es) and Turkish (tr).} (parallel texts and similar texts of the same topics in all languages), while the L2 part contains data from the same participants of different native languages reading 12 English texts (91 sentences, also encyclopedic-style). For each part, the complete texts were shown on multiple line on a single screen and the participants read naturally without any time limit. Furthermore, language-specific LexTALE tests have been carried out for several languages in the L1 experiments and the English version for all participants in the L2 experiment. LexTALE is a fast and efficient test of vocabulary knowledge for medium to highly proficient speakers \citep{lemhofer2012introducing}.

For comparison, we also run the experiments on the GECO corpus \citep{cop2017presenting}, which contains eye-tracking data from English and Dutch native speakers reading an entire novel in their native language (L1, 4921/4285 sentences, respectively), as well as a part where the Dutch speakers read English text (L2, 4521 sentences). The text was presented on the screen in paragraphs for natural unpaced reading.

\section{Results}
In the following, we show results for the correlation analysis across languages and an in-depth analysis on different influences on those correlations.

\paragraph{Languages}
We compute the Spearman correlation between relative fixation and first layer attention per sentence and average across sentences for all individual participants. We show correlation values averaged across participants for each language (L1) and corresponding data for English L2 in Table \ref{tab:lang}. We can see considerable differences between the languages, particularly in L1 with higher correlation values, e.g., for mBERT ($>0.5$) for \textit{et, fi, el, ru} and lower values ($<0.4$) for \textit{nl, en, it}. Correlations for XLM-R are about $0.1$ higher and for mT5 $0.1-0.2$ lower compared to mBERT. The correlation for English L2 are very similar between languages (0.3-0.34, mBERT) and lowest for the English L1 participants ($0.26$, mBERT). Correlation values for GECO are slightly lower for the Dutch experiments but in the same range for the English part.

\paragraph{Processing depth}
\begin{figure}[h!]
    \centering
    \includegraphics[width=.5\textwidth]{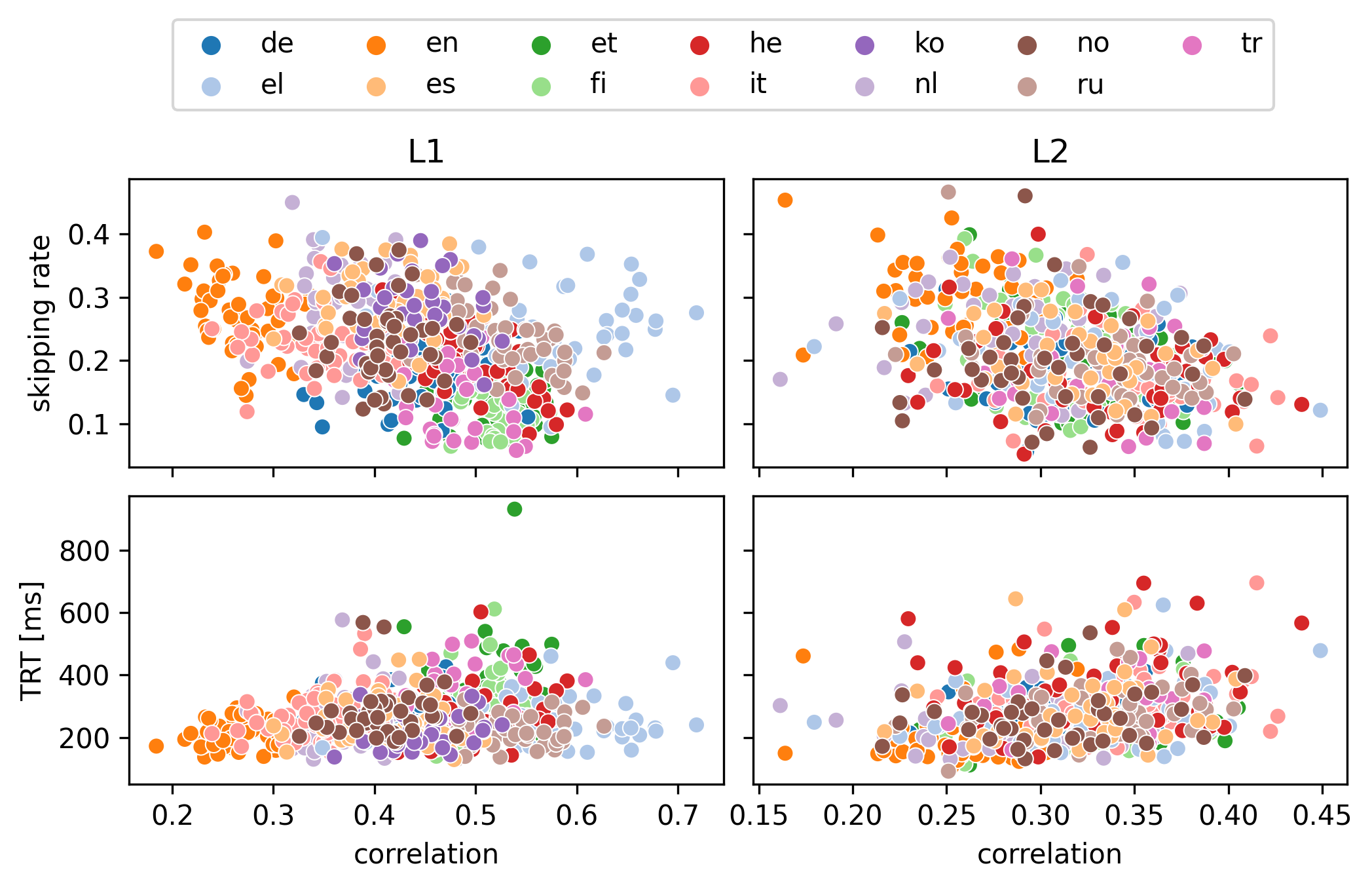}
    \caption{Correlation values for individual participants versus skipping rate (upper) and total reading time (lower) for L1 (left) and L2 (right) data. Spearman correlation was calculated on sentence-level and then averaged. Results are shown for mBERT.}
    \label{fig:skipping}
\end{figure}

To further analyse the different correlation values, particularly the low correlation in the L2 experiment for English native speakers, we look into skipping rates and total reading times and hereby focus on mBERT to make results more comparable to \citet{eberle-etal-2022-transformer}. Analyses on mT5 and XLM-R show similar results. Figure \ref{fig:skipping} shows skipping rates and total reading times computed for individual participants on the entire dataset versus individual correlation values as computed above. We find significant correlations ($p<0.01$) for both skipping rate vs.~correlation values ($-0.41/-0.34$) and TRT vs.~correlation values ($0.19/0.32$) for L1 and L2 respectively. This indicates that more thorough reading, i.e., less skipping and more time per word, leads to higher correlation with first layer attention. We also see those correlations at language-level for some languages where \textit{he, fi, ru} show highest scores at $-0.7, -0.63, -0.59$, respectively. For GECO, we find similar trends for English (L1 and L2) but not for Dutch.

\paragraph{POS}
\begin{figure}[h!]
\centering
\includegraphics[width=.4\textwidth]{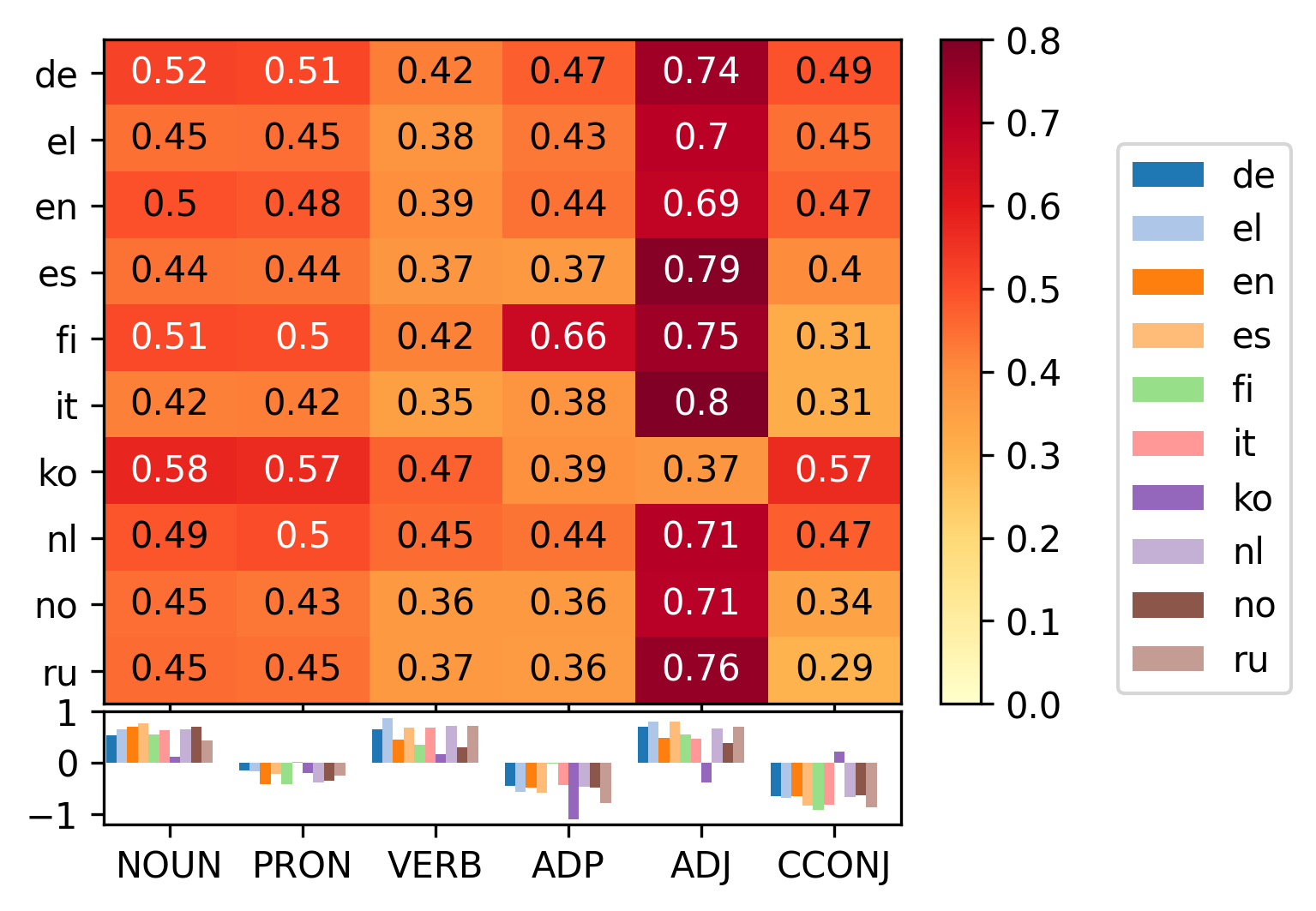}\\
\includegraphics[width=.4\textwidth]{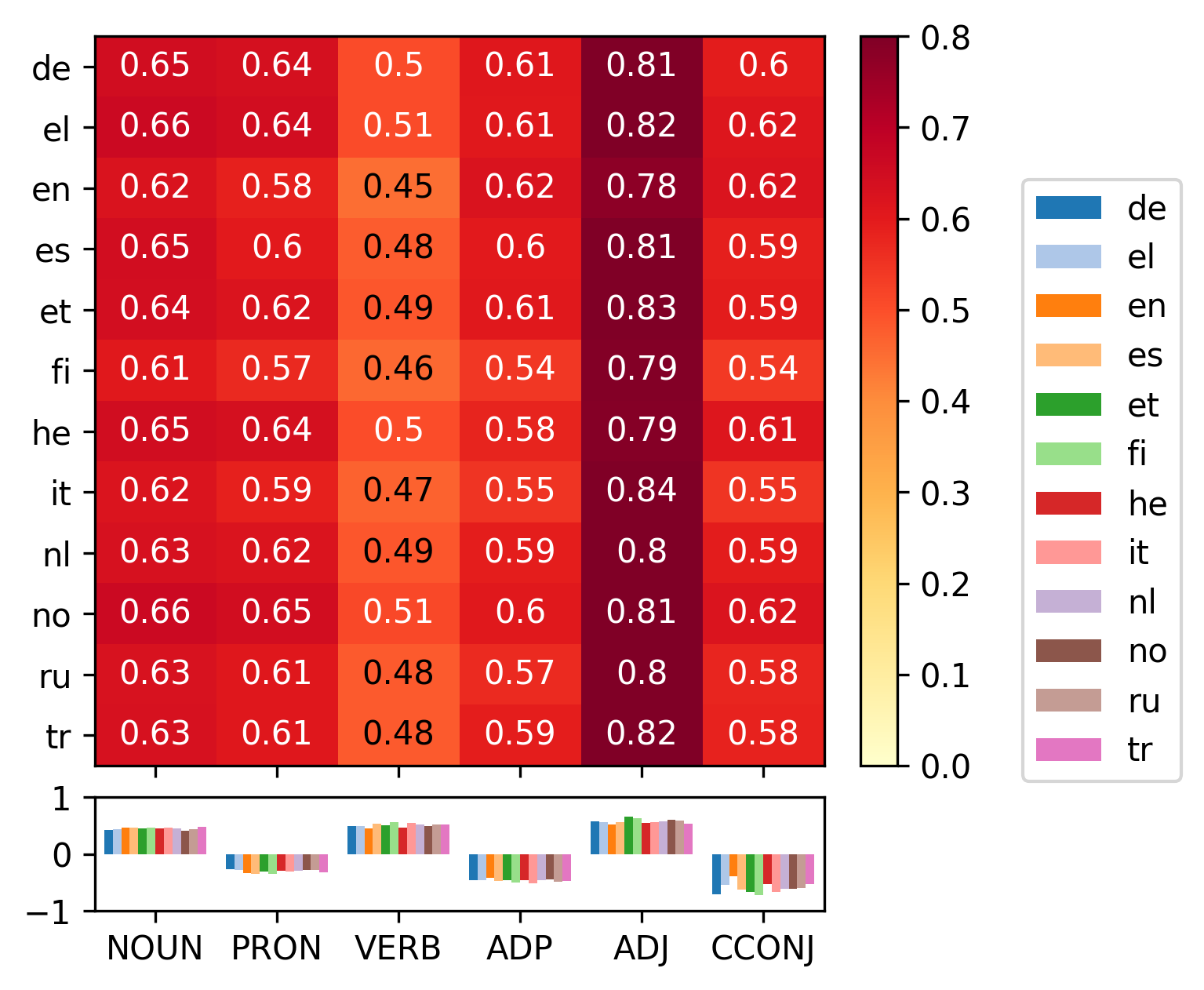}
\caption{Spearman correlations between human fixation and different languages for L1 and L2 for selected POS tags. Barplots show average attention value after standardisation (mean=0, std=1) for respective POS tag and model. For L1 only those languages are presented with an available POS-tagging model. Note that correlations are computed at token-level (not at sentence-level) which might cause higher correlations in L2. Results are shown for mBERT.}
\label{fig:pos}
\end{figure}

We look deeper into cross-lingual differences and show correlation values on token-level for 6 frequent POS tags in Figure \ref{fig:pos}. We extract relative fixations, standardise them to mean=0 and std=1 and average them across participants before computing the Spearman correlation with first layer attention values. We use POS-tagging models from \textit{spacy} and show results for the languages where respective models are available.\footnote{\url{https://spacy.io/usage/models}} Correlations for L1 are distributed similarly across different POS tags where adjectives show the highest correlation whereas verbs, although they carry an important part of the fixations, correlate much less. Only \textit{Korean} poses an exception here where adjectives do not play the most prominent role in human attention and also correlate much less. Here, nouns, pronouns, verbs and coordinating conjunctions correlate higher than in any other language and also much higher than adjectives. More research is required to interpret this finding. For L2, we see a very homogeneous distribution between languages and a similar distribution across POS tags as in most L1 experiments. 

\begin{figure}[h!]
    \centering
    \includegraphics[width=.44\textwidth]{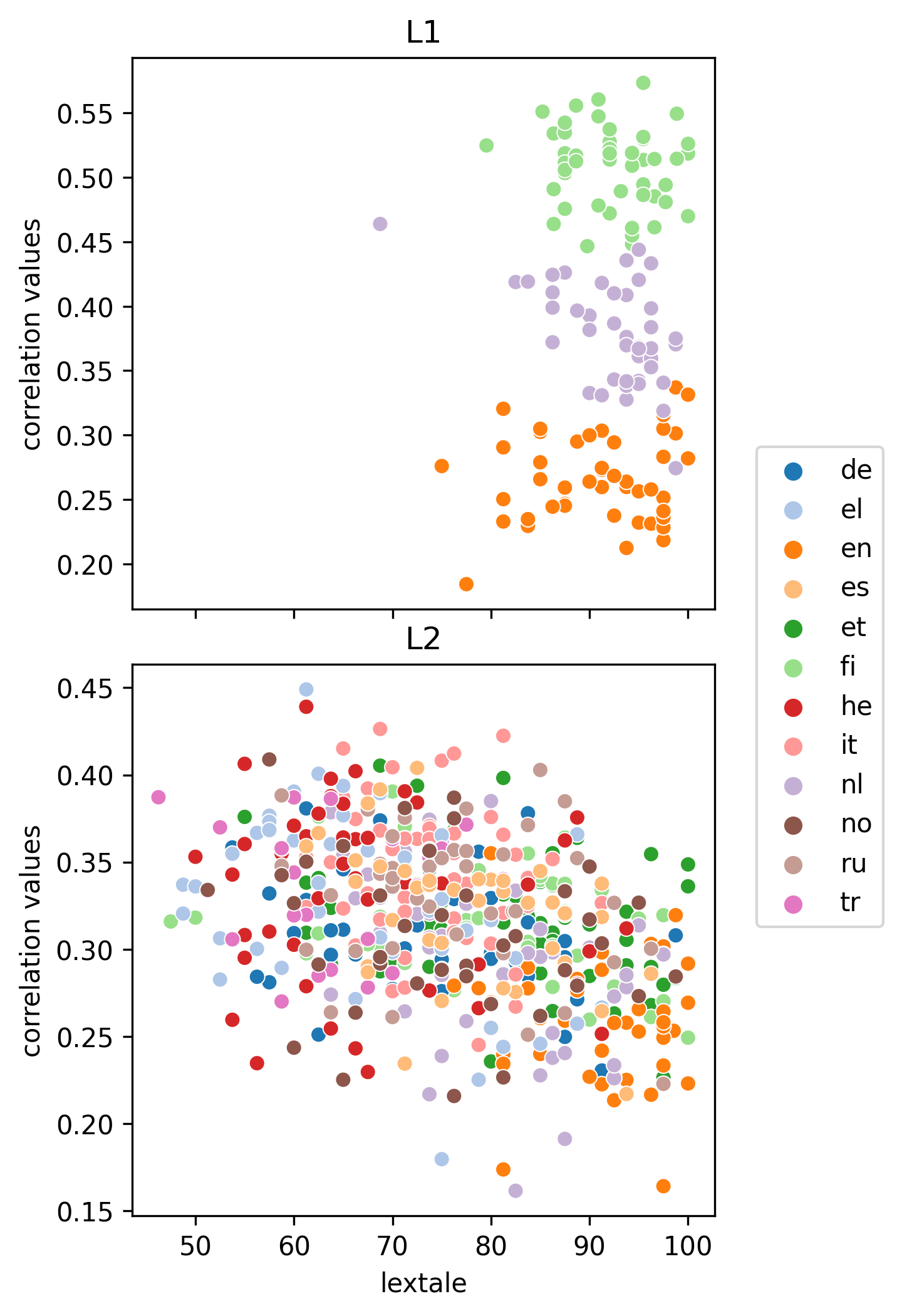}
    \caption{Spearman correlation values versus LexTALE score for individual participants for selected languages in L1 (\textit{en}, \textit{nl} and \textit{fi}) and all speakers in L2. Values for \textit{fi} in L1 were rescaled (with 100/88) to make them comparable. Results are shown for mBERT.}
    \label{fig:lextale}
\end{figure}
\paragraph{LexTALE}
 We show LexTALE scores for \textit{English} L2 and \textit{fi, en, nl} for L1 versus correlation values in Figure \ref{fig:lextale}. We find a negative correlation for Dutch speakers in L1 $-0.36$ and for the entire L2 data of $-0.42$ ($p<0.05$) suggesting that higher LexTALE scores lead to lower correlation with first layer attention.

\section{Discussion \& Conclusion}

Our results show that the correlation between relative fixation duration and first layer attention varies greatly across languages when read by native speakers. These differences can be attributed in part to the depth of processing: Languages such as Finnish and Greek, which show high total reading times, show a more evenly distributed correlation pattern across the most frequent parts of speech. Moreover, L1 English shows a high skipping rate and the lowest correlations. We find that more careful in-depth reading -- processing more words for a longer time -- correlates more strongly with attention than fast shallow reading. This is in line with previous research showing that attention patterns in BERT carry high entropy values, i.e., are broadly distributed, particularly in the first layers \cite{clark-etal-2019-bert}, which also leads to higher correlation with fixation duration \cite{eberle-etal-2022-transformer}.

The differences in skipping rate have various origins. On one hand, skipping rate is regulated by word length \citep{drieghe2004word}, which explains the lower skipping rate of agglutinative languages such as Finnish and Turkish \citep{meco_corpus}, and in turn their higher correlation to mBERT attention. On the other hand, word skipping is affected by L2 reading proficiency. More skilled learners make fewer fixations and skip more words \citep{dolgunsoz2016word}. This is reinforced by our comparison between English L2 and native English reading (which shows lower correlation).
This finding is also supported by our analysis on the LexTALE vocabulary test. LexTALE accurately estimates proficiency even at high levels \citep{ferre2017can}. Our results show that higher test scores lead to lower correlation with attention. Again, this is due to the reading depth: highly proficient readers have a higher skipping rate \citep{eskenazi2015reading}. 

We furthermore looked at the influence of age and gender but could not find any meaningful differences. This might be due to the fact that all participants were university students, most of them under the age of 30, thus representing a very specific group of the overall population. It is also important to note that most of the languages in MECO are Indo-European and only 4 are not using the Latin script. 

In summary, we have shown the impact of various subgroup characteristics reflected in reading and how they affect the correlation to neural attention. We argue that these differences should be taken into account when leveraging human language processing signals for NLP. 

\section*{Acknowledgements}
Stephanie Brandl was partially funded by the Platform Intelligence in News project, which is supported by Innovation Fund Denmark via the Grand Solutions program and from the European
Union’s Horizon 2020 research and innovation programme under the Marie Skłodowska-Curie grant
agreement No.~101065558. We thank Daniel Hershcovich for proof-reading and valuable inputs on the manuscript. 

\bibliography{anthology,custom}
\bibliographystyle{acl_natbib}




\end{document}